\documentclass{article}

\PassOptionsToPackage{numbers, compress}{natbib}
\usepackage[dblblindworkshop, final]{neurips_2026}

\workshoptitle{I Can't Believe It's Not Better (ICBINB): Failure Modes of AI in Biology}

\usepackage[utf8]{inputenc} % allow utf-8 input
\usepackage[T1]{fontenc}    % use 8-bit T1 fonts
\usepackage{hyperref}       % hyperlinks
\usepackage{url}            % simple URL typesetting
\usepackage{booktabs}       % professional-quality tables
\usepackage{amsfonts}       % blackboard math symbols
\usepackage{changes}        % for \added, \deleted, etc.
\usepackage{multirow}        % for multi-row cells in tables
\usepackage{graphicx}       % for \includegraphics
\usepackage{nicefrac}       % compact symbols for 1/2, etc.
\usepackage{microtype}      % microtypography
\usepackage{xcolor}         % colors
\usepackage{float}
\usepackage{wrapfig}
\usepackage{graphicx}
\usepackage{caption}
\usepackage{cuted}
\usepackage{needspace}

\newcommand{\pos}[1]{%
    \textcolor{green!50!black}{\scriptsize\,(+#1)}%
}
\newcommand{\negv}[1]{%
    \textcolor{red!75!black}{\scriptsize\,(-#1)}%
}
\newcommand{\same}[1]{%
    \textcolor{gray}{\scriptsize\,(#1)}%
}
\title{Open-Weight MLLMs Cannot Compare \\ What They Cannot Name}

\author{%
  Wenxin Su \\
  European Molecular Biology Laboratory\\
  Heidelberg, Germany\\
  \texttt{wenxin.su@embl.de} \\
  \And
  Anna Kreshuk \\
  European Molecular Biology Laboratory\\
  Heidelberg, Germany\\
  \texttt{anna.kreshuk@embl.de} 
}

\begin{document}

\maketitle

\begin{abstract}
  Multimodal Large Language Models (MLLMs) are increasingly capable of understanding images and performing complex visual reasoning, suggesting the possibility of using them to automate biological image analysis. This need is becoming increasingly urgent as advances in technology continue to generate ever larger and more diverse microscopy datasets, already exceeding the scale at which manual visual analysis is feasible. With the final aim to use MLLMs for visual comparison of complex phenotypes, we investigate a very simple baseline: comparison of the size of body parts in synthetic images of ants. We find that state-of-the-art open-weight MLLMs fail for some body parts, but not for others, even though the comparison itself is visually trivial for all of them. We evaluate different in-context learning strategies and analyze performance on the related object recognition task. Our results suggest a recognition-driven grounding failure: structures that the models cannot recognize are instead associated with a recognizable semantic concept, impairing the grounding and comparison of both. Further analysis of the vision encoder is consistent with the hypothesis that fine-grained visual structure may be less well preserved without semantic support. Together, our results reveal a surprising limitation of current MLLMs: semantic recognition can gate simple visual comparisons that do not intrinsically require semantic understanding. Our code and data are available at \url{https://github.com/kreshuklab/Open-Weight-MLLMs-Cannot-Compare-What-They-Cannot-Name}

\end{abstract}

\section{Problem}
\subsection{General setting}
Multimodal Large Language Models (MLLMs) combine text with other information modalities to enable solving cross-modal tasks such as visual question answering or image-based reasoning. Unlike LLMs that rely on external tools or models that combine independently pretrained unimodal representations only after unimodal training, MLLMs are trained to integrate multiple modalities within a shared architecture, typically through multimodal training objectives. Recent advances in open-weight MLLMs have narrowed the gap with proprietary frontier models, with leading open-weight models approaching frontier performance across a broad range of multimodal understanding and reasoning tasks \cite{yue2023mmmu, yue2025mmmu-pro}. Here, we use MLLM as the general term; since our experiments consider only vision and language, these models are also referred to as large vision-language models (LVLMs).

For biology, these models hold the promise of enabling microscopy data analysis through natural language interaction. Recent benchmarks have begun to probe this potential by evaluating microscopy-specific capabilities ranging from the recognition of imaging modalities and biological structures to object localization, fine-grained image interpretation, and research-level reasoning, such as hypothesis generation and experimental planning. \cite{burgess2025microvqa, lozano2024mubench, wei2026plantmicro}. 

Encouraged by the fast progress of MLLMs, we set out to explore their use for visual phenotype analysis. Such tasks are ubiquitous in biology and form the basis for the evaluation of perturbation effects from nano- to macro-scale. Scientists routinely compare the morphology of organelles, cells, organs and whole organisms, and importantly, such comparisons do not require an explicit understanding of the semantic identity of the objects being compared: humans can readily answer abstract questions such as “Is the top object larger in the left or right image?”. A single reference, or a small number of examples, is then often sufficient to associate an object with a semantic label. 

The ability to decouple visual comparison from prior semantic recognition is particularly important for biology. Mutations, pathological states, and experimental perturbations can produce morphologies that differ substantially from those represented in training data. At the same time, new microscopy modalities, staining protocols, and imaging techniques continually alter the visual appearance of otherwise familiar biological structures. A model limited to structures it already recognizes may therefore fail precisely in the experiments with greatest biological novelty. 
As vision models approach human-level performance in object recognition and scene understanding, it is natural to expect them to perform visually grounded comparisons as well, guided by questions expressed in natural language and, potentially, in a small number of visual examples. Such comparisons are especially attractive for automated phenotype analysis, where MLLMs could enable morphology-based measurements to scale to very large phenotypic screens and ultimately contribute to increasingly autonomous experimental workflows, from feedback microscopy to fully fledged AI co-scientists.

\begin{figure*}[t]
    \setlength{\belowcaptionskip}{-10pt}
    \begin{center}
        \includegraphics[width=0.98\linewidth,angle=0]{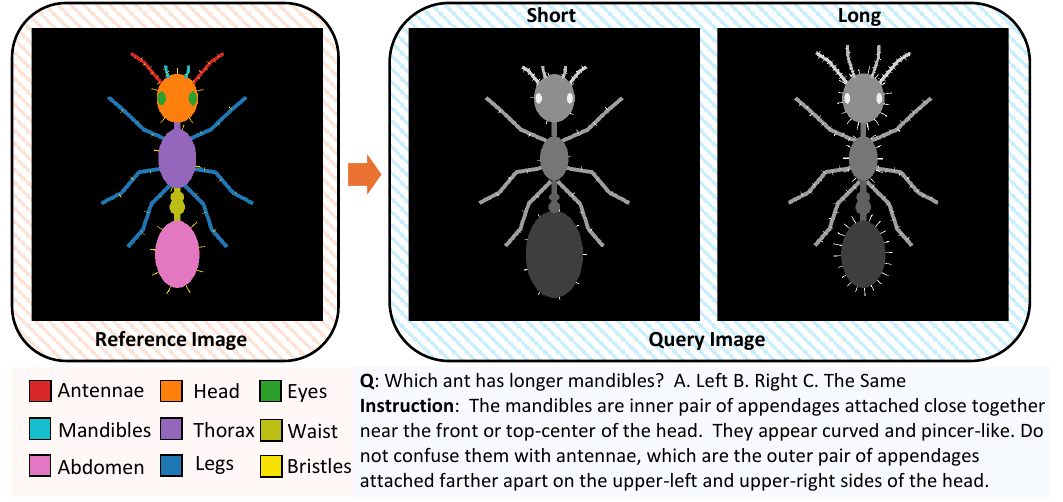}
    \end{center}
    \caption{
    Left: the ant and its 9 body parts that are  varied to generate the synthetic dataset. Right: the baseline visual comparison task with two ant images and the corresponding prompt.  
    } 
    \label{fig:ref} 
\end{figure*}

\begin{wrapfigure}{r}{0.44\textwidth}
\vspace{-28pt}
    \centering
    \includegraphics[width=1.0\linewidth]{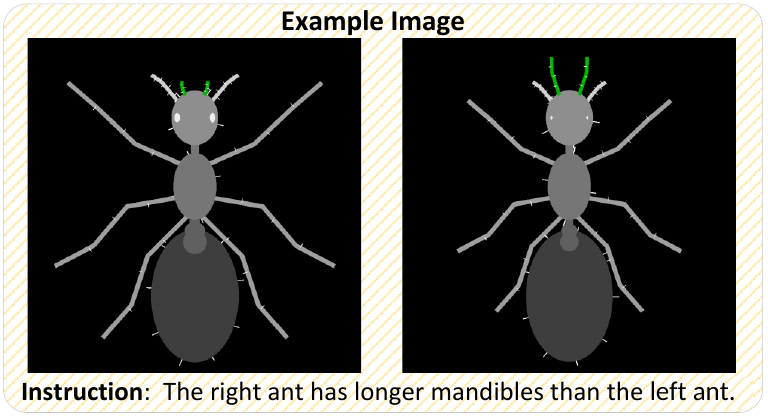}
    \caption{Example images provided for in-context learning. The target body part (mandibles) is highlighted in green, and the instruction indicates the correct answer.} 
    \label{fig:example}
\end{wrapfigure}

\subsection{Visual comparison task}
%A clearly specified biological task, data modality, and setting, with assumptions, target metrics, and desired improvements precisely stated.
%To simplify the setting and minimize the role of domain-specific knowledge, we focused on the whole-organism scale and posed questions requiring comparison of the sizes of anatomical features in insects. 
Our ultimate goal was to apply MLLMs to microscopy images generated in large-scale toxicity screens, such as those described in \cite{gandara2024pervasive}. The corresponding real microscopy dataset was available to us; however, before we proceeded to analyze the experimental data, we constructed a synthetic dataset to establish a baseline. Already in this simplified setting, we uncovered the failure mode that motivates the remainder of this work. We therefore demonstrate it using synthetic images of ants in dorsal view, in which the sizes of different anatomical parts are independently varied within specified ranges (see Fig.~\ref{fig:ref} left for an example, and \ref{sec_data_gen} for more details on data generation). 
The task was then to answer the question which of the two ants shown has a larger \{body part\}, where the {body part} could be one of \{legs, antennae, bristles, mandibles\} (Fig.~\ref{fig:ref} right). The question was formulated as multiple-choice, with possible answers "The left image has a larger \{body part\}", "The two images have approximately the same size \{body part\}", "The right image has a larger \{body part\}". The accuracy was measured simply as the percentage of correct answers in all the answers. The questions were deliberately made easy, with a large difference in the length of the target objects.

\section{Proposed approach}
%The modeling approach or solution under investigation, including its core mechanism, required preconditions, and the hypotheses under which it is expected to work.
We started from the prompt shown in Fig.~\ref{fig:ref} right: the two "experimental" images served as input to the models, but not the reference image. Beyond this baseline which is close to the ideal MLLM application setting, we evaluated several in-context learning approaches, providing the MLLMs with more and more guidance on the task, up to the full segmentation of the body part in question.
In more detail, we considered the following setups:
\begin{itemize}
    \item "Baseline": the full prompt is shown in Fig.~\ref{fig:prompt}.
    \item "Reference": the reference image of Fig.~\ref{fig:ref} left is provided along with the prompt.
    \item "Example": two ants with the segmented body part highlighted in green are provided as shown in Fig.~\ref{fig:example}. The query is for another pair without segmentation. 
    \item "Object": with the same example image of two ants, we refer to the body part as "object" in the query to avoid the semantic bias of biological prior knowledge.
    \item "Segmentation": in the query image the body part is segmented and highlighted in green.
    \item "Segmentation+Object": in the query image, we refer the segmented region as "object".   
\end{itemize}
\label{setup}

\section{Experiments and Observed Outcome} 
%A concise description of the negative, null, or unexpected outcome, with quantitative evidence where possible (metrics, uncertainty, compute budget, data/seed details).
\label{gen_inst}
\subsection{Data Generation}
\label{sec_data_gen}
%To our knowledge, there is no existing public biological datasets are designed to analyze the phenotype variation. Moreover, public datasets may overlap with the pretraining data of MLLMs, introducing potential data-contamination bias \cite{Mirage}. For the above reasons, we construct a synthetic dorsal-view ant structure dataset for controllable morphological analysis. 
We construct each sample by first placing the four main body parts, the head, thorax, waist, and abdomen, as the body axis. Attached structures including the eyes, legs, mandibles, antennae, and bristles are subsequently positioned relative to their corresponding main body components (Fig.~\ref{fig:ref}). We then generate morphological variants from this template by independently scaling each component around its fixed center. Each body part is assigned a predefined set of categorical levels, and we enumerate the Cartesian product of all attribute levels before randomly shuffling the resulting configurations. To introduce within-category variation, each categorical scale factor is multiplied by a Gaussian perturbation, $\epsilon \sim \mathcal{N}(1, 0.01^2)$. 
In total, we generated 20,000 images of ants and used them to randomly select 10,000 pairs for comparison, always maintaining a balance between "left", "same" and "right" answers.

% Then independently scale each part around its fixed center while maintaining connectivity between adjacent components. Attached structures including the eyes, legs, mandibles, antennae, and bristles are subsequently positioned relative to their corresponding body components. Their attachment points therefore adapt to changes in the body geometry, while their sizes or lengths are controlled independently. Each morphological attribute is sampled from a predefined set of categorical levels. Samples are generated from the Cartesian product of each body part's attribute levels and randomly shuffled. To introduce within-category variation, each categorical scale factor is multiplied by a Gaussian perturbation ($\epsilon \sim \mathcal{N}(1, 0.01^2)$).

% To construct the dataset, first it need to ensure the main body part's location center like head, thorax, waist, and abdomen. Then adjust body size with preserving center to ensure all the part is connected. and then for attached parts like eyes, legs, mandibles, antenna, and bristles, locate relative to the scaled body, and Independently change the attached part's size.

\subsection{Baseline prompt design}
Each prompt specifies the image setup, task goal, target body part location and shape, and comparison rules. For the baseline experiments, the full prompt is shown in Fig.~\ref{fig:prompt}. Similar prompts used for the in-context learning experiments are provided in the \texttt{Supplementary}.

\begin{figure}[htbp]
    \setlength{\belowcaptionskip}{-10pt}
    \begin{center}
        \includegraphics[width=0.98\linewidth,angle=0]{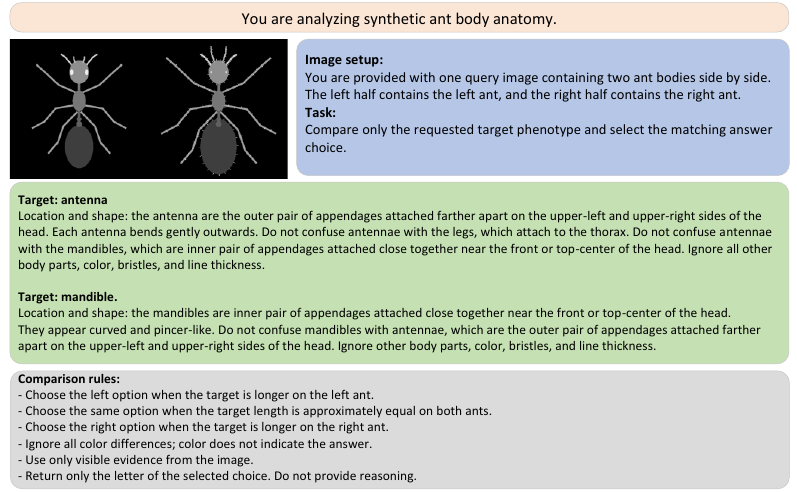}
    \end{center}
    \caption{
    Baseline visual comparison prompts for the "Antennae" and "Mandibles" classes.
    % for more reliable ViL prediction.
    } 
    \label{fig:prompt} 
\end{figure}

\begin{table}[t]
    \centering
    \caption{Baseline accuracy (\%) on size comparison tasks including "Antennae", "Mandibles", "Bristles", and "Legs" (random guessing: 33\%). Qwen3.8, Gemma4, and Qwen3.5 are evaluated in instruction and thinking modes. ``Param.'': number of parameters; ``Date'': model release date. Values in parentheses indicate the change relative to the instruction mode.}
    \label{tab:baseline}
    \small
    \setlength{\tabcolsep}{11pt}

    \begin{tabular}{@{}lcccccc@{}}
        \toprule
        {\textbf{Model}}
        & {\textbf{Param.}}
        & {\textbf{Date}}
        & \textbf{Antennae}
        & \textbf{Mandibles}
        & \textbf{Bristles}
        & \textbf{Legs}\\
        
        \midrule

        \multicolumn{7}{@{}l}{%
            \textbf{\textit{Open-weight Instruction MLLMs}}%
        } \\
        
        Qwen3.8
        & 27B & 2026-08 & 57.0 & 38.6 & \uline{\textbf{89.9}} & 90.5\\
        
        Gemma4
        & 31B & 2026-04 & 54.6 & 50.4 & 80.1 & \uline{\textbf{93.0}}\\
        
        Qwen3.5
        & 27B & 2026-02 & 51.9 & 33.2 & 83.7 & 83.8\\
        
        InternVL3.5
        & 38B & 2025-08 & 39.4 & 34.2 & 89.6 & 61.1\\
        
        GLM4.1
        & 9B & 2025-07 & 35.3 & 34.1 & 85.7 & 52.6\\
        
        \midrule

        \multicolumn{7}{@{}l}{%
            \textbf{\textit{Open-weight Thinking MLLMs}}%
        } \\
        
        Qwen3.8
        & 27B & 2026-08 &59.7\pos{2.7} & 56.0\pos{17.4} & 85.1\negv{4.8} & 90.3\negv{0.2}   \\
        
        Gemma4
        & 31B & 2026-04 & 50.4\negv{4.2} & 38.1\negv{12.3} & 77.9\negv{2.2} & 87.5\negv{5.5}\\
        
        Qwen3.5
        & 27B & 2026-02 & 50.4\negv{1.5} & 34.4\pos{1.2} & 83.2\negv{0.5} & 79.5\negv{4.3} \\
        \midrule
        ChatGPT5.6-Terra
        & -- & 2026-07 & \uline{\textbf{69.8}} & \uline{\textbf{79.7}} &83.4 &86.2   \\        
        \bottomrule
    \end{tabular}
\end{table}

\subsection{Model Selection}
We evaluate five current state-of-the-art open-weight models, Qwen3.8-27B \cite{Qwen38}, Qwen3.5-27B \cite{Qwen35}, Gemma4-31B \cite{team2026gemma}, InternVL3.5-38B \cite{wang2025internvl3}, and GLM4.1-9B \cite{hong2025glm}, in both instruction and thinking modes.  For clarity, we refer to each model using an abbreviated name consisting only of its model family and version; for example, Qwen3.8-27B is shortened to Qwen3.8. Table~\ref{tab:baseline} reports the parameter count and release date of each model. Due to computational constraints, all thinking-mode evaluations are conducted on a subset of 1,000 image pairs. We also evaluate the frontier proprietary model ChatGPT5.6-Terra \cite{Chatgpt56} on the same subset, from an API, without tool use.

\subsection{Baseline Result}
"Bristles" class appears to be easy for all models, while the models from 2026 succeed also in the "Legs" class. Visual comparison of the length attribute can thus be established as in general feasible for the open-weight models. Without the "Thinking" mode, the performance on the "Mandibles" class is close to random for all models except Gemma4, while for "Antennae" GLM4.1 is again close to random, but other models achieve much better results than for "Mandibles". Qwen3.8 is better than Qwen3.5 (38.6\% vs 33.2\%). With the "Thinking" mode on, Qwen3.8 improves significantly for the "Mandibles" class (from 38.6\% to 56.0\%), while Gemma4 gets worse and Qwen3.5 improves very slightly. "Thinking" mode consistently makes the comparison accuracy worse for the "Bristles" and "Legs" classes. ChatGPT5.6-Terra performs well on "Antennae" and better still on "Mandibles". In general, the results for "Antennae" and "Mandibles" are significantly worse than for "Legs" and "Bristles", motivating the introduction of in-context learning for "Antennae" and "Mandibles" classes. 

\subsection{In-Context Learning}
Addition of the reference image to the prompt brings a slight positive effect to the models from 2025 for both "Antennae" (Tab.~\ref{tab:antennae}) and "Mandibles" (Tab.~\ref{tab:mandibles}) classes. For Gemma4 and Qwen3.5 the reference image makes the performance worse, while for Qwen3.8 it's worse for "Antennae" and not affected for "Mandibles". 

Providing an example helps both Qwen models and InternVL3.5, and is detrimental for Gemma4 and GLM4.1. Interestingly, replacing the name of the body part by "Object" makes the performance of all models worse for the "Antennae" class, but InternVL3.5 and especially the Qwen models can benefit from it for the "Mandibles" class, allowing Qwen3.8 to overtake Gemma4 which was better at the baseline. An even larger performance boost is provided by the segmentation, where all models benefit at least to some extent. Most notably, Qwen3.8 and InternVL3.5 performance with segmentation is better for "Mandibles" than for "Antennae". 

Finally, the largest positive effect is provided by both including the segmentation and replacing the name of the body part by "Object". The only exception here is the GLM4.1 model. Here, all our attempts at in-context learning fail to significantly improve the accuracy compared to a random guessing baseline. All other models improve. Qwen3.5, also starting from a random guessing-level for the baseline task, achieves 76.8\% accuracy and overtakes Gemma4 which also improves substantially. While Qwen3.5 achieves the largest  improvement compared to the baseline for the "Mandibles" class, Qwen3.8 is the best open-weight model in absolute performance for both "Antennae" and "Mandibles", but, even with "Segmentation+Object" in-context guidance, it does not yet match its own baseline performance for "Legs" or "Bristles".  

\begin{table}[H]
    \centering
    \caption{Results (\%) on the \textbf{"Antennae"} task under different in-context learning setups. Baseline results are taken from Tab.~\ref{tab:baseline}. Values in parentheses indicate the change relative to the baseline.}
    \label{tab:antennae}
    \small
    \setlength{\tabcolsep}{6pt}

    \begin{tabular}{@{}lcccccc@{}}
        \toprule
        \textbf{Model}
        & \textbf{Baseline}
        & \textbf{Reference}
        & \textbf{Example}
        & \textbf{Object}
        & \textbf{Segmentation}
        & \textbf{Segmentation+Object} \\
        \midrule

        \multicolumn{7}{@{}l}{%
            \textbf{\textit{Open-weight Instruction MLLMs}}%
        } \\

        Qwen3.8
        & 57.0
        & 51.8\negv{5.2}
        & \uline{\textbf{60.0}}\pos{3.0}
        & 55.0\negv{2.0}
        & 75.7\pos{18.7}
        & \uline{\textbf{78.0}}\pos{21.0} \\

        Gemma4
        & 54.6
        & 48.0\negv{6.6}
        & 51.3\negv{3.3}
        & 49.9\negv{4.7}
        & 73.2\pos{18.6}
        & 75.9\pos{21.3} \\

        Qwen3.5
        & 51.9
        & 48.9\negv{3.0}
        & 53.8\pos{1.9}
        & 46.4\negv{5.5}
        & 72.0\pos{20.1}
        & 71.6\pos{19.7} \\

        InternVL3.5
        & 39.4
        & 42.5\pos{3.1}
        & 41.3\pos{1.9}
        & 34.5\negv{4.9}
        & 44.5\pos{5.1}
        & 43.7\pos{4.3} \\

        GLM4.1
        & 35.3
        & 35.5\pos{0.2}
        & 34.8\negv{0.5}
        & 33.4\negv{1.9}
        & 39.0\pos{3.7}
        & 34.9\negv{0.4} \\

        \midrule

        \multicolumn{7}{@{}l}{%
            \textbf{\textit{Open-weight Thinking MLLMs}}%
        } \\
        Qwen3.8 & \uline{\textbf{59.7}} & \uline{\textbf{54.5}}\negv{5.2} & 59.4\negv{0.3} &\uline{\textbf{59.0}}\negv{0.7} & \uline{\textbf{76.1}}\pos{16.4} & 73.2\pos{13.5} \\
        Gemma4  & 50.4 & 40.8\negv{9.6} & 51.2\pos{0.8} &50.9\pos{0.5} & 69.3\pos{18.9} &66.1\pos{15.7} \\
        Qwen3.5 & 50.4 & 49.2\negv{1.2} & 52.9\pos{2.5} & 47.9\negv{2.5} & 71.7\pos{21.3} & 70.3\pos{19.9} \\

        \bottomrule
    \end{tabular}
\end{table}

  % \vspace{-1.5em}

\begin{table}[H]
    \centering
    \caption{Results (\%) on the \textbf{"Mandibles"} task under different in-context learning setups. Baseline results are taken from Tab.~\ref{tab:baseline}. Values in parentheses indicate the change relative to the baseline.}
    \label{tab:mandibles}
    \small
    \setlength{\tabcolsep}{6pt}

    \begin{tabular}{@{}lcccccc@{}}
        \toprule
        {\textbf{Model}}
        & \textbf{Baseline}
        & \textbf{Reference}
        & \textbf{Example}
        & \textbf{Object}
        & \textbf{Segmentation}
        & \textbf{Segmentation+Object}\\
        \midrule

        \multicolumn{7}{@{}l}{%
            \textbf{\textit{Open-weight Instruction MLLMs}}%
        } \\

        Qwen3.8
        & 38.6
        & 38.6\same{0.0}
        & 40.7\pos{2.1}
        & 56.9\pos{18.3}
        & \uline{\textbf{77.9}}\pos{39.3}
        & \uline{\textbf{81.4}}\pos{42.8}\\

        Gemma4
        & 50.4
        & 44.1\negv{6.3}
        & 47.1\negv{3.3}
        & 48.5\negv{1.9}
        & 69.0\pos{18.6}
        & 73.0\pos{22.6}\\

        Qwen3.5
        & 33.2
        & 32.8\negv{0.4}
        & 34.3\pos{1.1}
        & 43.6\pos{10.4}
        & 68.8\pos{35.6}
        & 76.8\pos{43.6} \\

        InternVL3.5
        & 34.2
        & 35.4\pos{1.2}
        & 36.6\pos{2.4}
        & 36.6\pos{2.4}
        & 53.7\pos{19.5}
        & 51.6\pos{17.4}\\

        GLM4.1
        & 34.1
        & 34.2\pos{0.1}
        & 33.7\negv{0.4}
        & 34.1\same{0.0}
        & 34.3\pos{0.2}
        & 35.0\pos{0.9}\\

        \midrule

        \multicolumn{7}{@{}l}{%
            \textbf{\textit{Open-weight Thinking MLLMs}}%
        } \\

        Qwen3.8& \uline{\textbf{56.0}} & \uline{\textbf{55.9}}\negv{0.1} & \uline{\textbf{54.1}}\negv{1.9} &\uline{\textbf{59.2}}\pos{3.2} & 77.7\pos{21.7} & 77.4\pos{21.4} \\
        Gemma4  & 38.1 & 40.3\pos{2.1} & 34.7\negv{3.5}
        & 48.0\pos{9.9} & 70.7\pos{32.5} & 65.9\pos{27.7} \\
        Qwen3.5 & 34.4 & 32.1\negv{2.3} & 37.4\pos{3.0} & 42.3\pos{7.9} &69.9\pos{35.5} & 74.8\pos{40.4} \\

        \bottomrule
    \end{tabular}
\end{table}

%Based on these, we hypothesized that the failure is related to the model's ability to recognize the corresponding body part. The right half of Tab.~\ref{tab:evaluated_models} therefore shows object recognition task results, where the models were asked to name the body part highlighted in the query image (see Suppl for the full query). Further analysis of the errors on this task showed that antennae and mandibles are confused by the models. Even though antennae could be recognized correctly by all models, in the measurement task the mandibles were also identified as antennae which led to incorrect measurement for the antennae class. As mandibles themselves are almost never recognized correctly, the models also fail to measure them. Our in-context learning experiments therefore tried to help the models recognize the mandibles through examples.

\section{Reason for failure}
\subsection{Similar failures reported in related work}
It has previously been observed that MLLMs can struggle with relative visual judgments \cite{kil2024mllm,weng2025visnumbench}. However, the failure we observe differs from those reported in the benchmarks: we keep the comparison operation fixed and vary only the semantic label of visually similar target structures, resulting in dramatically different performance. For us, models can reliably compare the size of some anatomical structures while failing on others, suggesting that the problem lies not in the comparison operation itself but in associating the queried semantic concept with its visual representation. This interpretation is consistent with recent work documenting limitations in fine-grained visual attribute recognition and visual grounding in MLLMs \cite{yu2026fgbmk,villa2025merlim} although in these benchmarks "size" and "length" attributes are among the better performing ones. In \cite{yu2026fgbmk}, failures in fine-grained attribute recognition are hypothesized to stem from granularity inconsistencies between the image and text modalities. This is unlikely to be the root cause of our failure since the granularity is the same for all classes. Nor does the failure appear to reflect a general weakness in visual reasoning \cite{jiang2026blink}: the models successfully perform the same comparison for “Legs” and “Bristles”. Considering the stark improvement that the segmentation context brings for the newer models, we hypothesize that we are facing a more specific failure mode: the models struggle to ground certain semantic concepts in the visual representation required for the comparison. 
We investigate this hypothesis in two steps: (i) we evaluate how well the models assign semantic labels to body parts that have already been visually selected; and (ii) we examine the representations of the vision encoder.

\begin{center}
    % \centering

    % ---------- Left: Figure ----------
    \begin{minipage}[t]{0.43\textwidth}
        \vspace{0pt} % Force top alignment
        \centering

        \includegraphics[width=\linewidth]{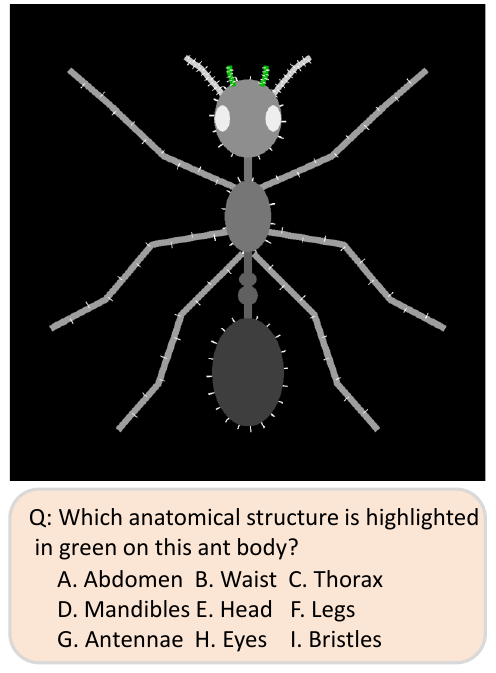}

        \captionof{figure}{
           Object recognition task setup.
        }
        \label{fig:object_task}
    \end{minipage}
    \hfill
    % ---------- Right: Table ----------
    \begin{minipage}[t]{0.54\textwidth}
        \vspace{0pt} % Force top alignment
        \centering

        \captionof{table}{
Accuracy (\%) on object recognition tasks of antennae and mandibles (random guessing: 11.1\%). Values in parentheses indicate the changes achieved with reference guidance relative to the baselines.
        }
        \label{tab:object_task}

        \scriptsize
        \setlength{\tabcolsep}{1.6pt}
        \renewcommand{\arraystretch}{1.30}

        \resizebox{\linewidth}{!}{%
        \begin{tabular}{@{}lcccc@{}}
            \toprule
            \multirow{2}{*}{\textbf{Model}}
            & \multicolumn{2}{c}{\textbf{Antennae}}
            & \multicolumn{2}{c}{\textbf{Mandibles}} \\

            \cmidrule(lr){2-3}
            \cmidrule(lr){4-5}

            & \textbf{Baseline}
            & \textbf{Reference}
            & \textbf{Baseline}
            & \textbf{Reference} \\
            \midrule

            \multicolumn{5}{@{}l}{%
                \textbf{\textit{Open-weight Instruction MLLMs}}%
            } \\

            Qwen3.8
            & 97.5 & 90.1\negv{7.4}
            & 0.4  & \uline{\textbf{31.8}}\pos{31.4} \\

            Gemma4
            & 96.9 & 99.0\pos{2.1}
            & \uline{\textbf{11.1}} & 3.4\negv{7.7} \\

            Qwen3.5
            & 99.7 & 99.6\negv{0.1}
            & 0.0  & 2.8\pos{2.8} \\

            InternVL3.5
            & \uline{\textbf{100.0}} & \uline{\textbf{100.0}}\same{0.0}
            & 0.0   & 0.7\pos{0.7} \\

            GLM4.1
            & \uline{\textbf{100.0}} & \uline{\textbf{100.0}}\same{0.0}
            & 0.0   & 0.0\same{0.0} \\

            \midrule

            \multicolumn{5}{@{}l}{%
                \textbf{\textit{Open-weight Thinking MLLMs}}%
            } \\

            Qwen3.8 & 98.2 & 96.7\negv{1.5} & 0.0 & 17.2\pos{17.2} \\
            Gemma4  & 97.3 & 97.6\pos{0.3} & 5.6 & 4.8\negv{0.8}\\
            Qwen3.5 & 99.9 & 99.7\negv{0.2} & 0.0 & 0.2\pos{0.2} \\

            \bottomrule
        \end{tabular}%
        }
    \end{minipage}

% \end{figure*}
% \end{strip}
\end{center}

\subsection{Object recognition}
For the failing “Antennae” and “Mandibles” classes, we assess the models’ ability to recognize the corresponding body parts in the whole-body image: the models are presented with an image of an ant in which the target body part is highlighted in green and are asked to identify the highlighted structure in a multiple-choice question (Fig.~\ref{fig:object_task}). We further evaluate a setting in which the model is provided with a reference image from Fig.~\ref{fig:ref}. The results are summarized in Tab.~\ref{tab:object_task}. 

The models recognize antennae with near perfect accuracy, but fail to recognize mandibles: Gemma4 achieves the random guessing accuracy of 11.1\%, other models are very consistently wrong for this class. Gemma4 performance gets worse when the reference image is provided. In contrast, the recent Qwen3.8 model benefits substantially from the reference image for recognition, improving from 0.4\% to 31.8\% (from 0\% to 17.2\% in Thinking mode), but this improvement does not carry over to the comparison task, for which the reference image has essentially no effect on performance. These results suggest that recognition ability may be necessary for visual comparison, but improving recognition through in-context learning does not necessarily improve performance for related comparison tasks. At the same time, Qwen3.8 in Thinking mode achieves 56\% accuracy on the baseline comparison task, but fails to recognize mandibles in the object recognition task, suggesting that some component in the Thinking mode chain-of-thought for this model can partially compensate for not being able to recognize. However, its 56\% accuracy is still far from the case when the semantic grounding is provided through the segmentation context.

For the "Antennae" class, near-perfect recognition also does not imply reliable visual comparison, with the best-performing model performance at 57\% compared to 93\% for the "legs" class. We hypothesize that this gap is at least partially explained by erroneous semantic grounding: while the models can correctly recognize antennae when they are highlighted, they mistake mandibles for antennae when they have to find antennae in the image themselves. We further analyze the distribution of wrong predictions on the mandible recognition task, and indeed, ~99\% of the misclassified mandibles are predicted as antennae.

\begin{figure*}[t]
    \setlength{\belowcaptionskip}{-10pt}
    \begin{center}
        \includegraphics[width=0.98\linewidth,angle=0]{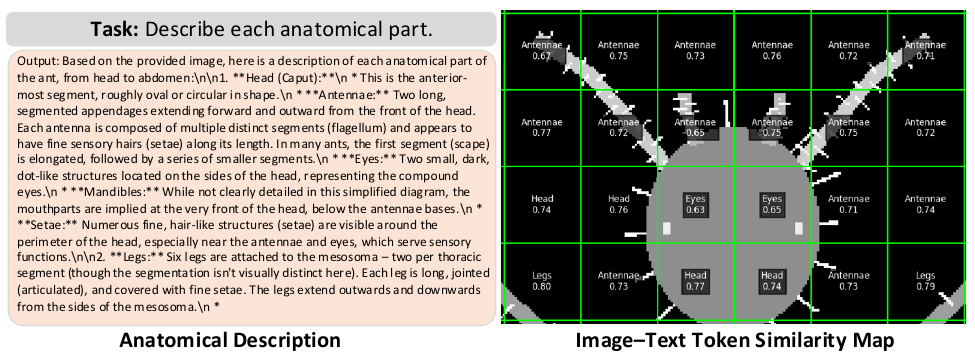}
    \end{center}
    \caption{
Left: anatomical description generated by Qwen3.8. Right: image–text token similarity map, where each image patch from the middle layer of the LLM decoder is assigned the body-part text token with the highest similarity.} 
    \label{fig:sim_text} 
\end{figure*}

We further investigate this hypothesis by directly visualizing each image patch's most strongly related body-part text token. First, we prompt Qwen3.8 to describe each body part, as shown in Fig.~\ref{fig:sim_text} Left. Then we compute the similarity between the image tokens from the middle layer of the MLLM decoder which encodes rich semantic information \cite{jiang2025devils} and body-part text tokens from the generated description. In Fig.~\ref{fig:sim_text} Right, each image patch shows the body-part text token with the highest similarity. Patches corresponding to the mandibles are most closely associated with the token "antennae", supporting our hypothesis.

% This hypothesis is supported by Fig.~\ref{fig:sim_text} which shows the body-part text token with the highest similarity for each image patch. Specifically,  We then assign to each image patch the body-part text token with the highest similarity, and patches corresponding to mandibles are most closely associated with the “antennae” token. 

Summarizing these results, we suggest two distinct failure modes for the visual comparison task: a recognition-driven failure, in which the target structure cannot be named when highlighted in the image, and a grounding-driven failure, in which a recognizable semantic concept is (additionally) associated with the wrong visual structures. In our case, the second is a consequence of the first, since the unrecognizable structures are present in the image and get wrongly associated with a recognizable class. In a different case, the unrecognizable structure might only cause its own comparison failure without affecting other classes. 

\begin{figure*}[t]
    \setlength{\belowcaptionskip}{-10pt}
    \begin{center}
        \includegraphics[width=0.98\linewidth,angle=0]{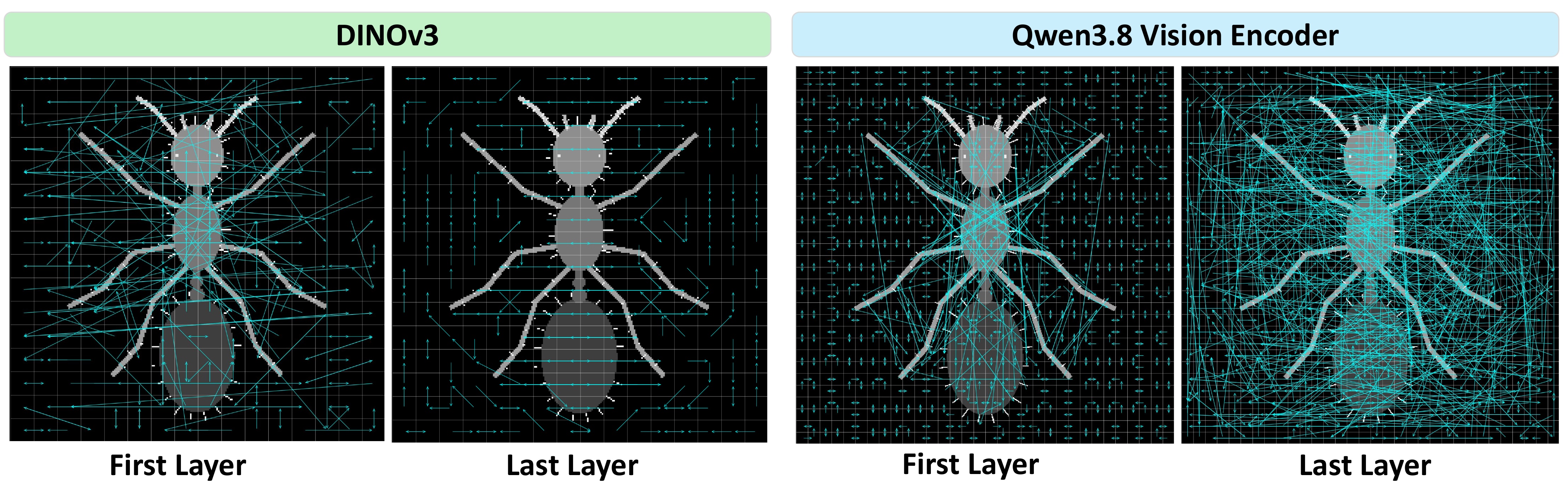}
    \end{center}
    \caption{
Top-1 most similar image token connections in the first and last layers of the DINOv3 and Qwen3.8 vision encoders. Blue arrows point from each patch to its most similar patch.
    } 
    \label{fig:sim_img} 
\end{figure*}

\subsection{Patch encoding analysis}
Next, we ask whether changes in visual representations associated with multimodal alignment could help explain the observed failure and its partial recovery under segmentation. This suggestion has also been made by \cite{yu2026fgbmk}, but while they find an effect on performance, they conclude that the granularity inconsistency matters more. In our case, the effect on performance is very strong, while the granularity inconsistency is unlikely to matter since it is the same for all classes, motivating a closer examination of the visual representations. We compare the first and last layers of the Qwen3.8 vision encoder with the layers of DINOv3 unimodal vision model \cite{simeoni2025dinov3} as shown in Fig.~\ref{fig:sim_img}. Within each image, we identify the most similar visual token for every token and connect the pairs, visualizing the relational structure of the patch space. For DINOv3, the first-layer features exhibit limited organization, whereas the final-layer features form a coherent structure aligned with semantic regions. For Qwen3.8, the first-layer features are more spatially organized than those of its final layer. This reversal suggests that multimodal alignment shifts the later layer representations toward compatibility with textual features, potentially weakening the within-image semantic structure preserved in earlier layers.

Here, we see a possible explanation for why comparison becomes difficult when the target cannot be semantically grounded: the model may be less able to rely on visual-feature similarity to group the relevant patches into a coherent object if the patches are not pulled together by the same text token.

\section{Conclusions}
We presented our attempts to perform visual attribute comparison in biological images using MLLMs. To isolate the core issue from other complexities of real experimental data, we constructed a synthetic dataset approximating whole-body insect imaging and evaluated MLLM performance on comparing the sizes of anatomical structures in ants. 
We find a strong association between models’ ability to recognize a target structure and their ability to reliably compare its visual attributes. Furthermore, failure to recognize one structure can distort the grounding of another, impairing comparisons even for concepts that the model can otherwise recognize. This is surprising because, for a human observer, the semantic identity of the structures is largely irrelevant to the comparison: estimating the object size is a simple perceptual operation independent of knowing what the object is. For the models, in-context learning improves performance, with the largest gains obtained if a segmentation of the structures to be compared is provided. While encouraging, this setting is not useful in practice, since measuring the size of segmented objects is a simple operation that does not require AI. We hypothesize that the failure arises from a loss of fine-grained visual structure in the vision encoder during multimodal training: visually similar image patches may cease to be represented similarly unless their similarity is reinforced through association with a shared semantic concept. Analysis of vision-text associations in the MLLM and of patch encodings in different layers of the Qwen3.8 vision encoder provides evidence consistent with this hypothesis. For open-weight MLLMs, semantic recognition thus appears to act as a gating mechanism for downstream visual reasoning, even for tasks that do not intrinsically require semantic understanding.

This limitation is particularly consequential for biological imaging, where the structures of interest may be altered by mutation, disease, perturbation, or new imaging conditions and therefore poorly represented by the model’s learned semantic associations. Still, while our results indicate that MLLMs cannot yet be reliably used for this type of visual reasoning, we see substantial grounds for optimism in the rapid improvement of model performance from 2025 to 2026, particularly in the progression from Qwen3.5 to Qwen3.8. The closed-source frontier model ChatGPT5.6-Terra leads the comparison for classes that open-weight models struggle to ground visually, whereas open-weight models outperform it on the more readily recognizable “Bristles” and “Legs” classes. We therefore hope that continued competition and progress at the frontier will pave the way to automated phenotype comparison and further visual reasoning in microscopy analysis pipelines.

\begin{ack}
This research was supported by EMBL IT for providing GPU computing resources.

\end{ack}

\bibliographystyle{unsrtnat}
\bibliography{references}

%%%%%%%%%%%%%%%%%%%%%%%%%%%%%%%%%%%%%%%%%%%%%%%%%%%%%%%%%%%%

\appendix

\section{Supplementary Material}

\subsection{Prompt Design for In-Context Learning}

\begin{figure}[htbp]
    \setlength{\belowcaptionskip}{-10pt}
    \begin{center}
        \includegraphics[width=0.98\linewidth,angle=0]{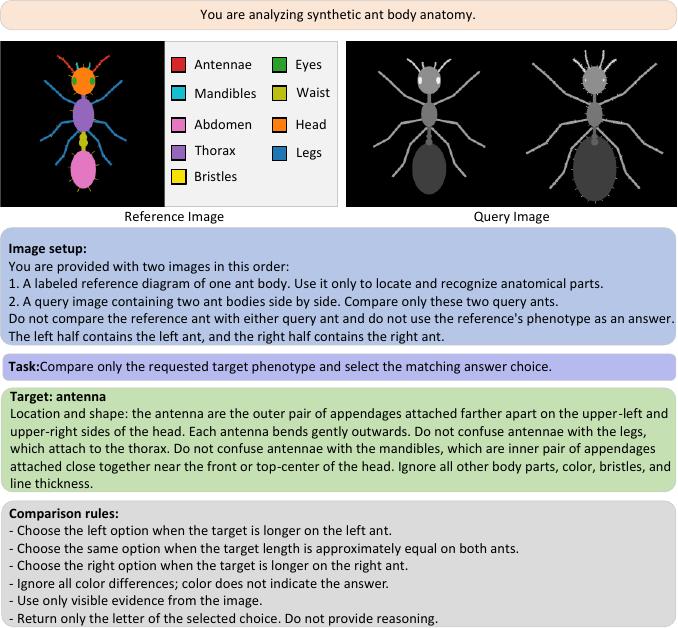}
    \end{center}
    \caption{
    Full prompt for the "Reference" setup of in-context learning.
    % for more reliable ViL prediction.
    } 
    \label{fig:prompt_ref} 
\end{figure}

\begin{figure*}[t]
    \setlength{\belowcaptionskip}{-10pt}
    \begin{center}
        \includegraphics[width=0.98\linewidth,angle=0]{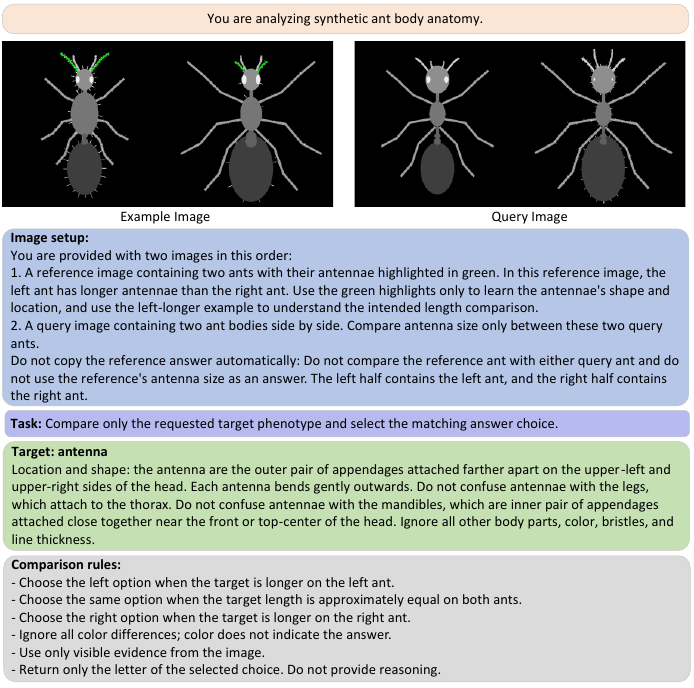}
    \end{center}
    \caption{
    Full prompt for the "Example" setup of in-context learning.
    % for more reliable ViL prediction.
    } 
    \label{fig:prompt_ex} 
\end{figure*}

\begin{figure*}[t]
    \setlength{\belowcaptionskip}{-10pt}
    \begin{center}
        \includegraphics[width=0.98\linewidth,angle=0]{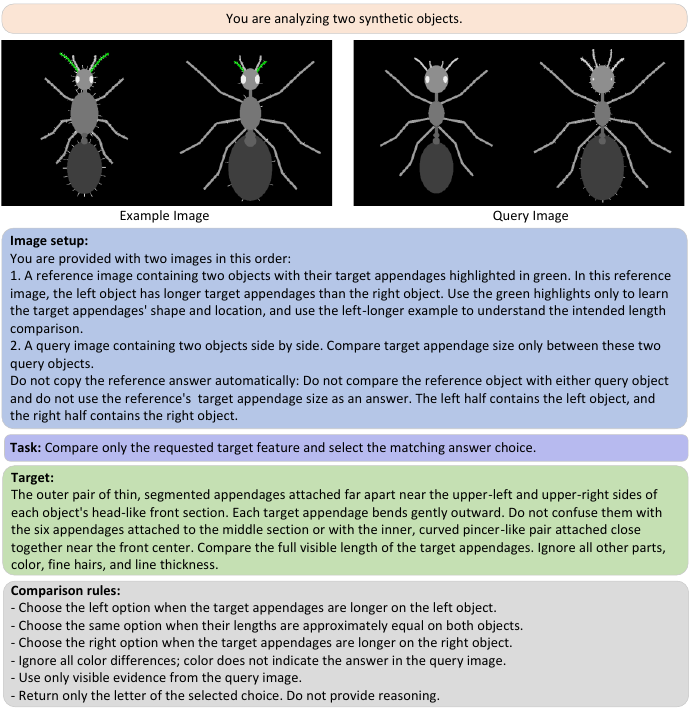}
    \end{center}
    \caption{
    Full prompt for the "Object" setup of in-context learning.
    % for more reliable ViL prediction.
    } 
    \label{fig:prompt_ob} 
\end{figure*}

\begin{figure*}[t]
    \setlength{\belowcaptionskip}{-10pt}
    \begin{center}
        \includegraphics[width=0.98\linewidth,angle=0]{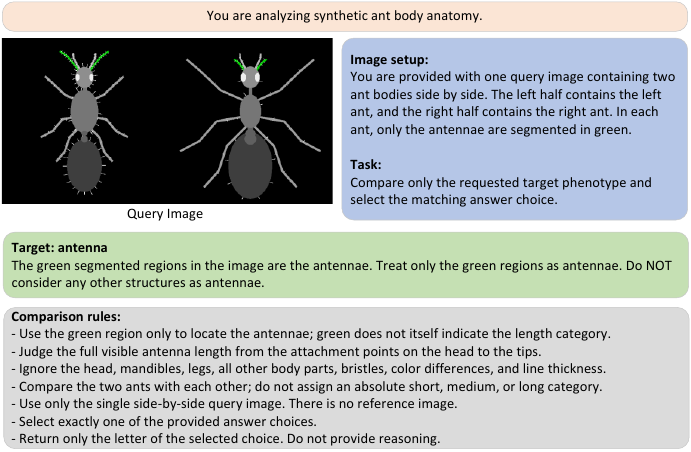}
    \end{center}
    \caption{
    Full prompt for the "Segmentation" setup of in-context learning.
    % for more reliable ViL prediction.
    } 
    \label{fig:prompt_seg} 
\end{figure*}

\begin{figure*}[t]
    \setlength{\belowcaptionskip}{-10pt}
    \begin{center}
        \includegraphics[width=0.98\linewidth,angle=0]{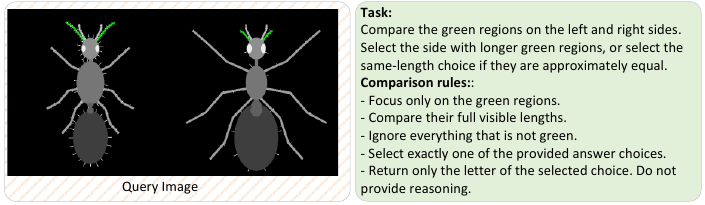}
    \end{center}
    \caption{
    Full prompt for the "Segmentation+Object" setup of in-context learning.
    % for more reliable ViL prediction.
    } 
    \label{fig:prompt_seg_ob} 
\end{figure*}

Figures~\ref{fig:prompt_ref}--\ref{fig:prompt_seg_ob} present the complete prompts used in the in-context learning setups of "Antennae" task. In the "Reference" setup, we provide a labeled anatomical diagram to help the MLLMs locate and recognize the target body part. In the "Example" setup, we provide an example in which the target body part is highlighted and the correct comparison is specified. In the "Object" setup, we replace biologically meaningful terms with generic expressions, referring to each ant as an object and the anatomical structure as the target appendage, thereby reducing reliance on biological semantic priors. In the "Segmentation" setup, we directly highlight the target body part in the query image, allowing the models to perform the comparison without first localizing it. Finally, the "Segmentation+Object" setup further removes biological terminology and instructs the models to directly compare the highlighted regions.

% \begin{figure*}[t]
%     \centering
%     \includegraphics[
%         width=0.98\textwidth,
%         pagebox=mediabox,
%         clip=false
%     ]{ob.pdf}
%     \caption{Full prompt for the ``Object'' setup of in-context learning.}
%     \label{fig:prompt_ob}
% \end{figure*}

% Technical appendices with additional results, figures, graphs, and proofs may be submitted with the paper submission before the full submission deadline (see above). You can upload a ZIP file for videos or code, but do not upload a separate PDF file for the appendix. There is no page limit for the technical appendices. 

% Note: Think of the appendix as ``optional reading'' for reviewers. The paper must be able to stand alone without the appendix; for example, adding critical experiments that support the main claims to an appendix is inappropriate. 

% \subsection{Further Patch Encoding Analysis}

%%%%%%%%%%%%%%%%%%%%%%%%%%%%%%%%%%%%%%%%%%%%%%%%%%%%%%%%%%%%

\end{document}